\documentclass[]{spie}  
\usepackage[]{graphicx}
\usepackage[]{amsmath}
\usepackage[]{amssymb}
\usepackage[]{caption}
\usepackage[]{algorithm2e}
\usepackage[]{subcaption} 
\usepackage[]{multirow}
\usepackage[]{array}

\DeclareMathOperator*{\dvg}{\operatorname{div}}

\title{Shape Complexes in Continuous Max-Flow Hierarchical Multi-Labeling Problems}

\author{John S.H. Baxter\supit{a,b}, Jing Yuan\supit{a,b}, and Terry M. Peters\supit{a,b}
\skiplinehalf
\supit{a}Robarts Research Institute, London, Ontario, Canada; \\
\supit{b}Western University, London, Ontario, Canada
}

\authorinfo{Send correspondence to J.S.H.B.: E-mail: jbaxter@robarts.ca}

\begin{document} 
  \maketitle 

\begin{abstract}
Although topological considerations amongst multiple labels have been previously investigated in the context of continuous max-flow image segmentation, similar investigations have yet to be made about shape considerations in a general and extendable manner. This paper presents \textit{shape complexes} for segmentation, which capture more complex shapes by combining multiple labels and super-labels constrained by \textit{geodesic star convexity}. Shape complexes combine geodesic star convexity constraints with hierarchical label organization, which together allow for more complex shapes to be represented. This framework avoids the use of co-ordinate system warping techniques to convert shape constraints into topological constraints, which may be ambiguous or ill-defined for certain segmentation problems.
\end{abstract}


\section{INTRODUCTION}
\label{sec:intro}
Since the proliferation of graph-cuts as an image segmentation mechanism\cite{boykov2001fast}, there has been a great amount of interest in expanding the methods to address problems in which more global information about the segmented objects is known. Topological considerations in terms of label orderings have allowed objects to be broken into and reassembled from distinct constituents, resulting in linear (or Ishikawa) models\cite{ishikawa2003exact}, submodular graph models\cite{delong2009globally}, and hierarchical models\cite{delong2012minimizing}. Global knowledge about the geometry of individual labels have led to approaches such as graph-cut-based shape templates\cite{vu2008shape} and graph-cut-accelerated statistical shape models\cite{malcolm2007graph}.

Shape information that can be implemented within a single graph-cut are particularly interesting in that they maintain submodularity and therefore guarantee global optimality. Often, these are formulated as constraints on the segmentation itself via additional infinite-cost edges between nodes. \textit{Simple star convexity}\cite{veksler2008star} and \textit{geodesic star convexity}\cite{gulshan2010geodesic} in particular use infinite-cost edges to constrain objects to be connected to a predefined vantage point through either a straight path (for simple star convexity) or a geodesic (for geodesic star convexity). 

For both label ordering considerations and shape information, continuous counterparts to discrete graph cuts have been developed. In terms of label ordering, Bae et al.\cite{bae2014fast} extended the Ishikawa model. Baxter et al.\cite{baxter2014ghmf,baxter2014dagmf} extended the hierarchical model and developed a framework for arbitrary label orderings. For simple and geodesic star convexity, Yuan et al.\cite{yuan2012efficient} and Ukwatta el al.\cite{ukwatta2013joint} provided two alternative solutions in the continuous domain. However, there has yet to be an general-purpose, extendable framework that combines topological and shape considerations simultaneously.

\section{Contributions}
The contribution of this paper is the definition of \textit{shape complexes}. Shape complexes are a collection of partially ordered labels in which several are constrained to have geodesic star convexity, thus addressing topological and shape considerations simultaneously. The result is that many additional shape constraints (such as rings) can be represented on multiple distinct objects simultaneously while maintaining a single co-ordinate system.

\section{Previous Work}
Yuan et al.\cite{yuan2012efficient} developed the first geodesic star convexity constraint in continuous max-flow segmentation, addressing optimization problems of the form:
\begin{equation}
\begin{aligned}
& \underset{u}{\min}
& & E(u) = \sum\limits_{\forall L} \int_\Omega \left( D(x)u(x)+S(x)|\nabla u(x)| \right) dx \\
& \text{subject to} & & u(x) \in  \{ 0, 1 \} \text{ and } \nabla u(x) \cdot e(x) \geq 0 \\
\end{aligned}
\end{equation}
where $u(x)$ is the labeling function and $e(x)$ is the local direction of the geodesic. This was addressed through the use of convex relaxation and primal-dual optimization using the labeling $u(x)$ and $1-u(x)$, as well as one additional field $\lambda(x)$ as multipliers. $\lambda(x)$ was used to ensure that the inequality constraint $\nabla u(x) \cdot e(x) \geq 0$ was respected, ensuring geodesic star convexity.

A very different approach was taken by Ukwatta et al.\cite{ukwatta2013joint} in order to implement geodesic star convexity. In his approach, the image co-ordinate system is warped, each in-plane image cross-section being represented in cylindrical co-ordinates with the center point placed somewhere in the object of interest, i.e. the interior blood pool of the femoral arterial. When expressed as such, simple star convexity can be expressed as a topological constraint, that is, as a continuous Ishikawa\cite{bae2014fast} model. This has the additional benefit that multiple simple star convexity constraints can be placed. In the framework proposed by Ukwatta et al.\cite{ukwatta2013joint}, both the vessel interior and the entire vessel (the union of the interior and wall labels) are constrained to have simple star convexity. Geodesic star convexity could be implemented by furthering warping of this co-ordinate system transformation.

\begin{figure}[h]
\centering
\begin{subfigure}{\textwidth}
\centering
\includegraphics[width=0.45\textwidth]{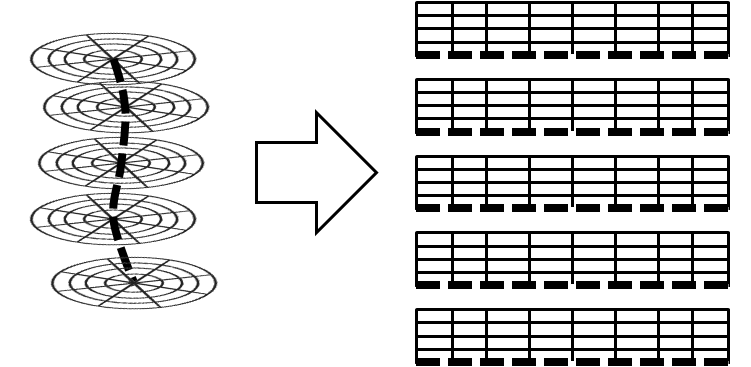}
\caption{Case in which cylindrical warping is neither ill-defined nor ambiguous.}
\end{subfigure} \\
\begin{subfigure}{0.49\textwidth}
\includegraphics[width=0.9\textwidth]{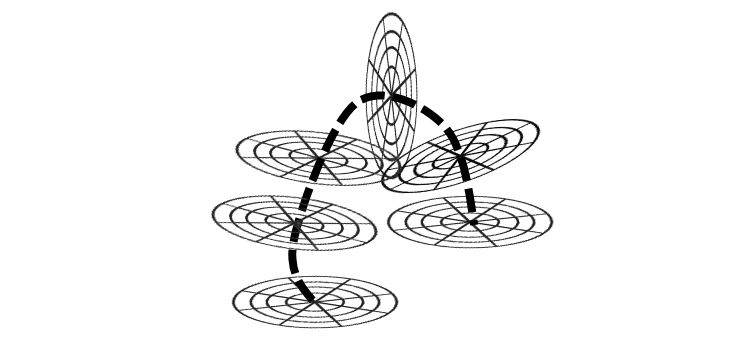}
\caption{Case in which cylindrical warping is ambiguous due to locations that are represented multiple times in the new warped co-ordinate system.}
\end{subfigure} \hspace{1mm}%
\begin{subfigure}{0.49\textwidth}
\includegraphics[width=0.9\textwidth]{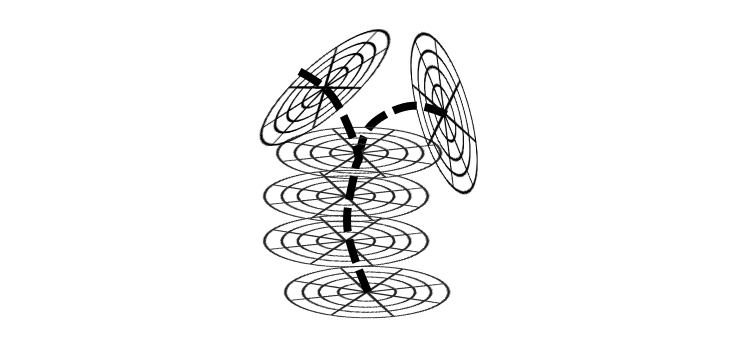}
\caption{Case in which cylindrical warping is ill-defined as no single curve can represent both distinct sets of center points (bifurcation).}
\end{subfigure}
\vspace{2mm}
\caption{Three possible cases for co-ordinate system warping}
\label{fig:warpingCases}
\end{figure}

However, these resampling approaches can be subject to ambiguity and ill-definition in a variety of cases. For the sake of example, consider vascular segmentation in which the vessel being segmentation bifurcates. After that bifurcation, no single center point can be defined, thus rendering the co-ordinate transformation ill-defined. The transformation may be ambiguous when the vascular `bends back' on itself creating a \textit{U} shape, causing multiple slices to intersect after warping. Illustrative versions of these three cases (including the non-ambiguous, well-defined case) are shown in Figure \ref{fig:warpingCases}.

Ultimately, the issues associated with co-ordinate system warping become more intractable as the number of distinct labels with geodesic star convexity increases, resulting in more ill-definition or warping ambiguity, as multiple interacting co-ordinate systems must be defined.

\section{Shape Complexes in Continuous Max-Flow Segmentation}
\subsection{Continuous Max-Flow Model}
This paper extends the work of Baxter et al.\cite{baxter2014ghmf} on \textit{generalized hierarchical continuous max-flow} (GHMF) models. Baxter et al. used the method of primal-dual optimization with augmented Lagrangian multipliers to address a max-flow min-cut problem through a directed tree of continuous spaces. The directed tree or hierarchy encodes the label ordering in the form of super-labels partitioned into their constituent parts with leaf nodes representing the ultimate partition of the whole image. The minimization equation we intend to solve extends the GHMF minimization equation:
\begin{equation}
\begin{aligned}
& \underset{u}{\min}
& & E(u) = \sum\limits_{\forall L} \int_\Omega \left( D_L(x)u_L(x)+S_{L}(x)|\nabla u_L(x)| \right) dx \\
& \text{subject to} & & \forall L (u_L(x) \geq 0) \\
& & &  \forall L \left(\sum\limits_{ L' \in L.C }u_{L'}(x) = u_L(x)\right) \\
& & &   u_S(x)=1
\end{aligned}
\label{dual_form}
\end{equation}
by the addition of the constraint originally posed by Yuan et al.\cite{yuan2012efficient}:
\begin{equation*}
\begin{aligned}
\nabla u_L(x) \cdot e_L(x) \geq 0
\end{aligned}
\label{shape_cons}
\end{equation*}
This constraint dictates that for each label at each pixel, there is a direction $e_L(x)$ in which assigning the label $L$ to voxel $x$ entails giving the same label to voxels proximal to $x$ in the positive $e_L(x)$ direction. If all of these direction vectors point at a single spatial location, this enforces simple star convexity. If this direction field is warped, it enforces geodesic star convexity.

The approach taken will be analogous to the augmented Lagrangian approach taken by Yuan et al.\cite{yuan2010study,yuan2012efficient}.

\subsection{Primal Formulation}
As in the previous work, the primal formulation will be expressed as a max-flow optimization, specifically:
\begin{equation}
\label{primal-first}
\begin{aligned}
\underset{p,q',\lambda} \max & \int_\Omega p_S(x) dx
\end{aligned}
\end{equation}
subject to the capacity constraints:
\begin{equation*}
\begin{aligned}
p_L(x) & \leq D_L(x), \text{ } L \in \text{ leaves}\\
|q'_L(x)| & \leq S_L(x) \text{ } L \neq S \\
\lambda_L(x) & \geq 0
\end{aligned}
\end{equation*}
and the flow conservation constraint:
\begin{equation*}
0 =\operatorname{div} \left( q'_L(x) + \lambda_L(x) e_L(x) \right) + p_L(x) - p_{L.P}(x) \text{ .}
\end{equation*}
$q'_L(x)$, the spatial flow not including flow along the infinite capacity direction $e_L(x)$, is distinguished from $q_L(x)$, the overall spatial flow, noting that $q_L(x) = q'_L(x) + \lambda_L(x)e_L(x)$. Here we depart from the approach taken by Yuan et al.\cite{yuan2012efficient} by rewriting Eq. \eqref{primal-first} in terms of $q_L(x)$ instead of $q'_L(x)$ yielding the formula:
\begin{equation*}
\label{primal-rewrite}
\begin{aligned}
\underset{p,q,\lambda} \max & \int_\Omega p_S(x) dx
\end{aligned}
\end{equation*}
subject to the constraints:
\begin{equation*}
\begin{aligned}
p_L(x) & \leq D_L(x), \text{ } L \in \text{ leaves}\\
|q_L(x)-\lambda_L(x)e_L(x)| & \leq S_L(x) \text{ } L \neq S \\
\lambda_L(x) & \geq 0 \\
0 & =\operatorname{div} q_L(x) + p_L(x) - p_{L.P}(x)
\end{aligned}
\end{equation*}
in which it should be noted that $\lambda$ can be explicitly solved for, appearing only in the constraints $\lambda(x) \geq 0$ and $|q_L(x)-\lambda_L(x)e_L(x)| \leq S_L(x)$. In order to ensure the spatial capacity constraint, the value of  $|q_L(x)-\lambda_L(x)e_L(x)|$ should be minimized over $\lambda_L(x) \geq 0$. This can be done analytically by finding the projection of $q_L(x)$ onto $e_L(x)$, ignoring any component of $q_L(x)$ in the $-e_L(x)$ direction. Thus, $\lambda_L(x) = \max \{ 0, q_L(x) \cdot e_L(x) \}$ assuming without loss of generality that $e_L(x)$ is unit length if a geodesic direction is specified and zero length if not. The final primal formulation is:
\begin{equation}
\label{primal}
\begin{aligned}
\underset{p,q} \max & \int_\Omega p_S(x) dx
\end{aligned}
\end{equation}
subject to the constraints:
\begin{equation*}
\begin{aligned}
p_L(x) & \leq D_L(x), \text{ } L \in \text{ leaves}\\
|q_L(x)-\lambda(x)e_L(x)| & \leq S_L(x) \text{ } L \neq S \text{ where } \lambda_L(x) = \max \{ 0, q_L(x) \cdot e_L(x) \} \\
0 & =\operatorname{div} q_L(x) + p_L(x) - p_{L.P}(x)
\end{aligned}
\end{equation*}

\subsection{Primal-Dual Formulation}
In order to develop the primal-dual formulation, we use the method of Lagrangian multipliers over the primal equation \eqref{primal} on the flow conservation constraint  $G_L(x) = \operatorname{div} q_L(x) + p_L(x) - p_{L.P}(x) = 0$, yielding the equation:
\begin{equation}
\label{lagrangian}
\begin{aligned}
\underset{u} \min \underset{p,q} \max & \left( \int_\Omega p_S(x) dx + \sum_{\forall L \neq S} \int_\Omega u_L(x) G_L(x) dx \right) \\
p_L(x) & \leq D_L(x), \text{ } L \in \text{ leaves}\\
|q_L(x)-\lambda(x)e_L(x)| & \leq S_L(x) \text{ } L \neq S \text{ where } \lambda_L(x) = \max \{ 0, q_L(x) \cdot e_L(x) \} \text{ .}
\end{aligned}
\end{equation} 

To ensure that this equation is optimizable, meeting the requirements of the minimax theorem, we must ensure that it is convex with respect to $u$, considering $p,q$ to be fixed, and concave with respect to $p,q$ with $u$ fixed.\cite{ekeland1976convex} Holding $p$ and $q$ fixed, $G$ is also fixed and thus Eq. \eqref{lagrangian} is linear with no constraints and is therefore concave. If $u$ is fixed, Eq. \eqref{lagrangian} is linear with convex constraints and is therefore convex. Thus  Eq. \eqref{lagrangian} can be optimized.

\subsection{Dual Formulation}
To show the equivalence of Eqs. \eqref{dual_form} and \eqref{lagrangian}, we will consider the optimization of each set of flows in a bottom-up manner similar to GHMF\cite{baxter2014ghmf}. Starting with any leaf node in the tree we can isolate $p_L$ in \eqref{lagrangian} as:
\begin{equation}
\underset{u} \min \underset{p_L(x) \leq D_L(x)} \max \int_\Omega u_L(x)p_L(x)dx = \underset{u, u_L(x) \geq 0} \min  \int_\Omega u_L(x)D_L(x)dx
\end{equation}
(If $u_L(x) < 0$, the function can be arbitrarily maximized by $p_L(x) \to -\infty$.) Every branch label can then be isolated as:
\begin{equation}
\underset{u} \min \underset{p_L(x)} \max \left( \int_\Omega u_L(x)p_L(x)dx - \sum_{\forall L' \in L.C} \int_\Omega u_{L'}(x)p_L(x)dx \right) = 0
\end{equation}
at the saddle point defined by $u_L(x) =  \sum_{\forall L' \in L.C} u_{L'}(x)$. The source flow, $p_S$, can be isolated similarly as:
\begin{equation}
\underset{u} \min \underset{p_S(x)} \max \left( \int_\Omega p_S(x)dx - \sum_{\forall L' \in S.C} \int_\Omega u_{L'}(x)p_S(x)dx \right) = 0
\end{equation}
at the saddle point defined by $1 =  \sum_{\forall L' \in S.C} u_{L'}(x)$. These constraints combined yield the labeling constraints in the original formulation.

To maximize the spatial flows, we will use the first primal formulation Eq. \eqref{primal-first} and separate the spatial flows into $q'_L(x)$ and $\lambda_L(x)e_L(x)$. Considering $q'_L(x)$ first, the maximization can be expressed as described by Giusti \cite{giusti1984minimal}] as:
\begin{equation} \label{eq:regularSpatialFlow}
\begin{aligned}
\int_\Omega u_L(x) \dvg q'_L(x)dx
& = \int_\Omega \left( \dvg (u_L(x)q'_L(x)) - q'(x) \cdot \nabla u_L(x) \right) dx \\
& = \int_\Omega \dvg (u_L(x)q'_L(x)) dx  - \int_\Omega  q'_L(x) \cdot \nabla u_L(x) dx \\
& = \oint_{\delta\Omega} u_L(x)q'_L(x) \cdot d\mathbf{s}  - \int_\Omega  q'_L(x) \cdot \nabla u_L(x) dx \\
& = - \int_\Omega  q'_L(x) \cdot \nabla u_L(x) dx \\
\underset{|q'_L| \leq S_L(x)} \max \int_\Omega u_L(x) \dvg q'_L(x)dx
& = \underset{|q'_L| \leq S_L(x)} \max  - \int_\Omega  q'_L(x) \cdot \nabla u_L(x) dx \\
& =  - \int_\Omega  \left( -\frac{S_L(x)}{|\nabla u_L(x)|} \nabla u_L(x) \right) \cdot \nabla u_L(x) dx \\
& = \int_\Omega S_L(x) |\nabla u_L(x)| dx \\
\underset{u_L \geq 0} \min \; \; \underset{|q'_L| \leq S_L(x)} \max \int_\Omega u_L(x) \dvg q_L(x)dx 
& = \underset{u_L \geq 0} \min \int_\Omega S_L(x) |\nabla u_L(x)| dx \text{ .}
\end{aligned}
\end{equation}

Taking the same approach to $\lambda_L(x)e_L(x)$:
\begin{equation} \label{eq:exemptionSpatialFlow}
\begin{aligned}
\int_\Omega u_L(x) \dvg \left( \lambda_L(x)e_L(x) \right) dx
& = \int_\Omega u_L(x) \left( \nabla \lambda(x) \cdot e_L(x) \right) dx + \int_\Omega u_L(x) \lambda(x) \dvg e_L(x) dx \\
& = \int_\Omega \nabla \lambda(x) \cdot u_L(x)e_L(x) dx + \int_\Omega \lambda(x) \dvg \left( u_L(x) e_L(x) \right) dx -  \int_\Omega \lambda(x) \left( \nabla u_L(x) \cdot e_L(x) \right) \\
& = \int_\Omega \dvg \left( \lambda(x) u_L(x) e_L(x) \right) dx -  \int_\Omega \lambda(x) \left( \nabla u_L(x) \cdot e_L(x) \right) \\
& = \oint_{\delta \Omega} \lambda(x) u_L(x) e_L(x) \cdot d\mathbf{s} -  \int_\Omega \lambda(x) \left( \nabla u_L(x) \cdot e_L(x) \right) \\
& = -  \int_\Omega \lambda(x) \left( \nabla u_L(x) \cdot e_L(x) \right)
\end{aligned}
\end{equation}
which can be arbitrarily maximized if $\nabla u_L(x) \cdot e_L(x) < 0$ by allowing $\lambda(x) \to \infty$. Thus, $\lambda(x)$ acts as a multiplier over the constraint $\nabla u_L(x) \cdot e_L(x) \geq 0$ and ensures the geodesic star convexity constraint on Eq. \eqref{dual_form}.

Thus, the primal Eq. \eqref{primal}, primal-dual Eq. \eqref{lagrangian}, and dual Eq. \eqref{dual_form} are all equivalent.

\subsection{Pseudo-Flow Model}
An alternative approach can be found by taking advantage of the pseudo-flow framework for GHMF explored by Baxter et al.\cite{baxter2015proximal} Ultimately, the framework is equivalent with the exception of the Chambolle iteration in which the exemption flow must be removed prior to the projection operation and subsequently added back in. Thus, the spatial flow update step becomes:
\begin{equation}
\begin{aligned}
q_L(x) \gets & \left\{ \begin{array}{ll}
q_L(x) - c\tau \nabla \left( v_L(x) \exp \left( - \frac{d_L(x)}{c} \right) \right) , & \text{ if } L \in \mathbb{L} \\
q_L(x) - c\tau \nabla \left( \sum_{L' \in \mathbb{L}} W_{(L,L')} v_{L'}(x) \exp \left( - \frac{d_L'(x)}{c} \right)\right) & \text{ else }
\end{array} \right. \\
\lambda(x) \gets & \max \{ 0, \, q_L(x) \cdot e_L(x) \\
q_L(x) \gets & q_L(x) - \lambda(x)e_L(x) \\
q_L(x) \gets & \lambda(x)e_L(x) + \operatorname{Proj}_{|q_L(x)|\leq S_L(x)} q_L
\end{aligned}
\end{equation}

\section{Solution to Primal-Dual Formulation}
To address the optimization problem, we first augment Eq. \eqref{lagrangian} to ensure convergence towards feasible solutions\cite{bertsekas1999nonlinear}:
\begin{equation}
\label{augmented}
\begin{aligned}
\underset{u} \min \underset{p,q} \max & \left( \int_\Omega p_S(x) dx + \sum_{\forall L \neq S} \int_\Omega u_L(x) G_L(x) dx - \frac{c}{2} \sum_{\forall L \neq S} \int_\Omega G_L(x)^2 dx \right)\\
p_L(x) & \leq D_L(x), \text{ } L \in \text{ leaves}\\
|q_L(x)-\lambda_L(x)e_L(x)| & \leq S_L(x) \text{ } L \neq S
\end{aligned}
\end{equation}
where $c > 0$ is an additional positive parameter penalizing deviation from the flow conservation constraint. 

Eq. \eqref{augmented} can be optimized using each variable independently. We use the following steps iteratively:
{
\begin{enumerate}
\setlength{\belowdisplayskip}{2pt} \setlength{\belowdisplayshortskip}{2pt}
\setlength{\abovedisplayskip}{2pt} \setlength{\abovedisplayshortskip}{2pt}
\item Maximize \eqref{augmented} over $q_L$ at all levels which is a Chambolle projection iteration\cite{chambolle2004algorithm} with descent parameter $\tau > 0$ by the following steps:
\begin{equation*}
\begin{aligned}
q_L(x) \gets & q_L(x) + \tau \nabla \left( \dvg q_L(x) + p_L(x) - p_{L.P}(x) - u_L(x)/c \right) \\
\lambda(x) \gets & \max\{ 0, \; q_L(x) \cdot e_L(x) \} \\
q'_L(x) \gets & q_L(x) - \lambda_L(x)e_L(x) \\
q'_L(x) \gets & \operatorname{Proj}_{|q'_L(x)| \leq S_L(x) } q'_L(x)  \\
q_L(x) \gets & q'_L(x) + \lambda_L(x)e_L(x) \\
\end{aligned}
\end{equation*}
The calculation, subtraction, and subsequent addition of  $\lambda_L(x)e_L(x)$ allows for the descent to occur over the entire spatial flow $q_L(x)$ but the spatial capacity to be applied only to the flow in directions not exempted by the infinite-capacity direction $e_L(x)$.
\item Maximize \eqref{augmented} over $p_L$ at the leaves analytically by:
\begin{equation*}
p_L(x) \gets \min\lbrace D_L(x), p_{L.P}(x) - \dvg q_L(x) + u_L(x)/c \rbrace
\end{equation*}
\item Maximize \eqref{augmented} over $p_L$ at the branches analytically by:
\begin{equation*}
p_L(x) \gets \frac{1}{|L.C|+1} \left( (p_{L.P}(x) - \dvg q_L(x) + u_L(x)/c) + \sum_{\forall L' \in L.C}\left( p_{L'}(x) + \dvg q_{L'}(x) - u_{L'}(x)/c \right) \right)
\end{equation*}
\item Maximize \eqref{augmented} over $p_S$ analytically by:
\begin{equation*}
p_S(x) \gets \frac{1}{|S.C|} \left( 1/c + \sum_{\forall L' \in S.C}\left( p_{L'}(x) + \dvg q_{L'}(x) - u_{L'}(x)/c \right) \right)
\end{equation*}
\item Minimize \eqref{augmented} over $u_L$ analytically by:
\begin{equation*}
u_L(x) \gets u_L(x) - c \left( \dvg q_L(x) - p_{L.P}(x) + p_L(x) \right)
\end{equation*}
\end{enumerate}
}

\subsection{Generalized Hierarchical Max-Flow Algorithm with Geodesic Star Convexity}
\subsubsection{Augmented Lagrangian Approach}
The algorithm used is an extension of the GHMF algorithm\cite{baxter2014ghmf}, taking advantage of its general framework for label ordering. The key difference between algorithms is the use of $\lambda$ and $e_L$ to calculate how much of the spatial flow $q_L$ is exempted from the spatial capacity constraint $|q_L(x)| \leq S_L(x)$. This exempted flow is removed prior to the projection operation and added back in after said operation. The sequential algorithm used is: \newpage
\begin{algorithm}[!h]
InitializeSolution(S) \;
\While{not converged} {
  \For{ $\forall L \neq S$ } {
    $\forall x, q_L(x) \gets q_L + \tau \nabla \left( \dvg q_L(x) + p_L(x) - p_{L.P}(x) - u_L(x)/c \right)$ \;
    $\forall x, \lambda(x) \gets \max \{ 0, q_L(x) \cdot e_L(x) \}$ \;
    $\forall x, q_L(x) \gets q_L - \lambda(x) e_L(x)$ \;
    $\forall x, q_L(x) \gets \operatorname{Proj}_{|q_L(x)| \leq S_L(x) } \left( q_L \right)$ \;
    $\forall x, q_L(x) \gets q_L + \lambda(x) e_L(x)$ \;
  }
  UpdateSinkFlows($S$) \;
  \For{ $\forall L \neq S$ } {
    $\forall x, u_L(x) \gets u_L(x) - c \left( \dvg q_L(x) - p_{L.P}(x) + p_L(x) \right)$ \;
  }
}
\label{alg:solverSeq}
\end{algorithm}

\noindent which makes use of the function:

\begin{algorithm}[!h]
{ \bf UpdateSinkFlows(L) } \\
\For{ $\forall L' \in L.C$} {
  UpdateSinkFlows($L'$) \;
}
\If{ $L$ is a leaf }{
  $\forall x, p_L(x) \gets \min\lbrace D_L(x), p_{L.P}(x) - \dvg q_L(x) + u_L(x)/c \rbrace$ \;
}
\If{ $L = S$ }{
  $\forall x, p_S(x) \gets 1/c$ \;
  \For{ $\forall L' \in S.C$ } {
     $\forall x, p_S(x) \gets p_S(x) + p_{L'}(x) + \dvg q_{L'}(x) - u_{L'}(x)/c $ \;
  }
  $\forall x, p_S(x) \gets \frac{1}{|S.C|}p_S(x) $ \;
}
\If{ $L$ is a branch }{
  $\forall x, p_L(x) \gets p_{L.P}(x) - \dvg q_L(x) + u_L(x)/c $ \;
  \For{ $\forall L' \in L.C$ } {
     $\forall x, p_L(x) \gets p_L(x) + p_{L'}(x) + \dvg q_{L'}(x) - u_{L'}(x)/c$ \;
  }
  $\forall x, p_L(x) \gets \frac{1}{|L.C|+1} p_L(x)$ \;
}
\end{algorithm}
For the sake of conciseness, the `{\bf for} \textit{$\forall x$} {\bf do} ' loops surrounding each assignment operation have been replaced with the prefix $\forall x$ in both the algorithm and the function definitions. The initialization mechanism, InitializeSolution(S), is flexible and the current algorithm initializes as $\forall L,$ $u_L(x)=0$. Alternative approaches could include the original GHMF initialization using the minimum cost solution with no regularization or shape constraints.

\newpage
\subsubsection{Pseudo-Flow Approach}
The pseudo-flow approach can be made into the following algorithm, using the notation from Baxter et al.\cite{baxter2015proximal}:\\
\begin{algorithm}[h!]
$\forall L \in \mathbb{L}, \, u_L(x) \gets 1 /  |\mathbb{L}| $\;
\While{not converged}{
  $\forall L, \, d_L(x) \gets \dvg q_L(x)$\;
  $\forall L \in \mathbb{L}, \, d_L(x) \gets d_L(x) + D_L(x)$\;
  \For{$L$ in order $\mathbb{O}/ \{ S \} $}{
  	\For{$L' \in L.C$}{
  		$d_{L'}(x) \gets d_{L'}(x) + d_L(x)$\;
  	}
  }
  
  $\forall L \in \mathbb{L}, \, u_L(x) \gets u_L(x) \exp \left( - \frac{d_L(x) }{c} \right) $\;
  $\forall L \in \mathbb{L}, \, d_L(x) \gets u_L(x)$\;
  $a(x) \gets \sum_{\forall L\in \mathbb{L}}u_L(x)$\;
  $\forall L\in \mathbb{L}, \, u_L(x) \gets u_L(x) / a(x)$\;
  
  $\forall L \notin \mathbb{L}, \, d_L(x) \gets 0$\;
  \For{$L$ in order $\mathbb{O}^{-1} / \{ S \}$}{
  	
  	$q_L(x) \gets q_L(x) - c\tau \nabla d_L(x) $ \;
  	$\lambda(x) \gets \max \{ 0, \, q_L(x) \cdot e_L(x) \}$ \;
  	$q_L(x) \gets q_L(x) - \lambda(x)e_L(x)$ \;
  	$q_L(x) \gets \lambda(x)e_L(x) + \operatorname{Proj}_{|q_L(x)| \leq S_L(x)}  q_L(x)$ \;
  	\For{$L' \in L.P /  \{ S \} $}{
		$d_{L'}(x) \gets d_{L'} + d_L(x)$\;
	}
  }
}
\end{algorithm}

\noindent where $\mathbb{L}$ is the set of all leaf labels (where $L.C = \emptyset$). $\mathbb{O}$ and $\mathbb{O}^{-1}$ are top-down and bottom-up orderings through the hierarchy respectively.

\section{Discussion and Conclusions}
In this paper, we present an algorithm for addressing continuous max-flow problems in which shape complexes, collections of ordered labels with geodesic star convexity constraints, are defined. Shape complexes allow for a variety of more complex shapes to be represented, such as rings, as well as multiple distinct complexes without possibly ill-defined or ambiguous co-ordinate warping transformations.

In terms of future work, there are many different directions in which to take shape complexes. From an implementation point of view, the next steps will be the incorporation of shape complexes into more flexible label ordering models such as directed acyclic graph max-flow models\cite{baxter2014dagmf} which would increase the flexibility of the framework both in terms of possible models for regularization.

\acknowledgments 
The authors would like to acknowledge Dr.\ Jiro Inoue, Dr.\ Elvis Chen and Jonathan McLeod for their invaluable discussion, editing, and technical support.


\bibliographystyle{spiebib}   
\bibliography{TechReportShapeComplex} 

\end{document}